\documentclass{article}
\usepackage[latin9]{inputenc}
\usepackage{verbatim}
\usepackage{mathtools}
\usepackage{amsthm}
\usepackage{amsmath}
\usepackage{amssymb}
\usepackage{graphicx}
\usepackage{esint}
\usepackage[unicode=true,pdfusetitle,
 bookmarks=true,bookmarksnumbered=false,bookmarksopen=false,
 breaklinks=false,pdfborder={0 0 1},backref=section,colorlinks=false]
 {hyperref}
\hypersetup{
 plainpages=false,bookmarksdepth=-2}
\usepackage{breakurl}

\makeatletter

\newcommand{\noun}[1]{\textsc{#1}}
\providecommand{\tabularnewline}{\\}

\theoremstyle{plain}
\newtheorem{thm}{\protect\theoremname}
  \theoremstyle{plain}
  \newtheorem{lem}[thm]{\protect\lemmaname}
  \theoremstyle{plain}
  \newtheorem{cor}[thm]{\protect\corollaryname}

\usepackage{proceed2e}

\usepackage[flushleft]{threeparttable}

\newcommand{\half}[1]{\frac{#1}{2}}

\usepackage{tikz}\usetikzlibrary{calc}
\usetikzlibrary{svg.path}

\newcommand*{\defeq}{\mathrel{\rlap{%
                     \raisebox{0.3ex}{$\m@th\cdot$}}%
                     \raisebox{-0.3ex}{$\m@th\cdot$}}%
                     =}

\newcommand{\smallbullet}{\, \begin{picture}(-1,1)(-1,-3)\circle*{3}\end{picture}\,\, }

\usepackage{enumitem}\setlist{nolistsep}
\newcommand{\overbar}[1]{\mkern 1.5mu\overline{\mkern-1.5mu#1\mkern-1.5mu}\mkern 1.5mu}

\global\long\def\argmax{\operatornamewithlimits{argmax}}

\usepackage{relsize}\def\ssum{\operatornamewithlimits{\mathsmaller{\sum}}}
\usepackage{scrextend}
\setenumerate[0]{leftmargin=5pt,itemindent=5pt,labelwidth=5pt}

\global\long\def\half#1{\frac{#1}{2}}
\global\long\def\bd#1{\boldsymbol{#1}}

\title{Factorized Asymptotic Bayesian Inference for Factorial Hidden Markov
Models}

\author{ \textbf{Shaohua Li} \\
 Nanyang Technological University \\ Singapore \\
 \texttt{shaohua@gmail.com} \\
 \And
 \textbf{Ryohei Fujimaki} \\
 NEC Laboratories America \\
 \texttt{rfujimaki@nec-labs.com} \\
 \And
 \textbf{Chunyan Miao} \\
 Nanyang Technological University \\ Singapore \\
 \texttt{ascymiao@ntu.edu.sg}
}

\makeatother

  \providecommand{\corollaryname}{Corollary}
  \providecommand{\lemmaname}{Lemma}
\providecommand{\theoremname}{Theorem}

\begin{document}
\maketitle 
\begin{abstract}
Factorial hidden Markov models (FHMMs) are powerful tools of modeling
sequential data. Learning FHMMs yields a challenging simultaneous
model selection issue, i.e., selecting the number of multiple Markov
chains and the dimensionality of each chain. Our main contribution
is to address this model selection issue by extending Factorized Asymptotic
Bayesian (FAB) inference to FHMMs. First, we offer a better approximation
of marginal log-likelihood than the previous FAB inference. Our key
idea is to integrate out transition probabilities, yet still apply
the Laplace approximation to emission probabilities. Second, {\normalsize{}we
prove that if there are two very similar hidden states in an FHMM,
i.e. one is redundant, then FAB} will almost surely shrink and eliminate
one of them, making the model parsimonious. Experimental results show
that {\normalsize{}FAB for FHMMs }significantly outperforms state-of-the-art
nonparametric Bayesian iFHMM and Variational FHMM in model selection
accuracy, with competitive held-out perplexity. 
\end{abstract}

\section{Introduction}

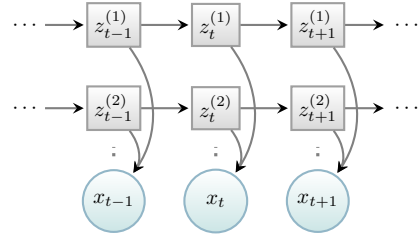
\begin{figure}[htbp]
\setlength{\abovecaptionskip}{0pt}\setlength{\belowcaptionskip}{-10pt}\begin{centering}

\begin{centering}
\begin{tikzpicture}[scale=0.8, every node/.style={scale=0.8},   point/.style={coordinate},   >=stealth,thick,draw=black!50,   tip/.style={->,shorten >=1pt},   hv path/.style={to path={-| (\tikztotarget)}},   vh path/.style={to path={|- (\tikztotarget)}},   state/.style={rectangle, minimum height=7mm, minimum width=7mm, draw=gray!50!black!50,   top color=white, bottom color=gray!50!black!20,    font=\normalfont},   obs/.style={circle,   minimum height=6mm, minimum width=6mm, draw=cyan!50!black!50,   top color=white, bottom color=cyan!50!black!20,  font=\normalfont},   ]   \matrix[row sep=0.5cm,column sep=0.5cm,ampersand replacement=\&]{       \node (fanin1) {\(\cdots\)};     \& \node (z11) [state]  {\(z_{t-1}^{(1)}\)};     \& \node (z12) [state]  {\(z_t^{(1)}\)};     \& \node (z13)   [state]  {\(z_{t+1}^{(1)}\)  };     \& \node (fanout1) {\(\cdots\)};\\     \node (fanin2) {\(\cdots\)};     \& \node (z21) [state]  {\(z_{t-1}^{(2)}\)};     \& \node (z22) [state]  {\(z_t^{(2)}\)};     \& \node (z23)   [state]  {\(z_{t+1}^{(2)}\)  };     \& \node (fanout2) {\(\cdots\)};\\     \node {};      \& \node (x1) [obs]    {\(x_{t-1}\)};     \& \node (x2) [obs]    {\(\;\;   x_t \;\;  \)};     \& \node (x3)   [obs]    {\(x_{t+1}\)  };     \& \node {}; \\     };   \path 	(fanin1) edge[tip] (z11)         (z11)   edge[tip] (z12)         (z12)   edge[tip] (z13) 	(z13) edge[tip] (fanout1) 	(fanin2) edge[tip] (z21)         (z21)   edge[tip] (z22)         (z22)   edge[tip] (z23) 	(z23) edge[tip] (fanout2) 	(z11) edge[bend left=35,tip] (x1) 	(z21) edge[bend left=35,tip] (x1) 	(z12) edge[bend left=35,tip] (x2) 	(z22) edge[bend left=35,tip] (x2) 	(z13) edge[bend left=35,tip] (x3) 	(z23) edge[bend left=35,tip] (x3) 	($(z21)+(0,-0.8)$) edge[dotted,very thick] ($(x1)+(0,0.9)$) 	($(z22)+(0,-0.8)$) edge[dotted,very thick] ($(x2)+(0,0.9)$) 	($(z23)+(0,-0.8)$) edge[dotted,very thick] ($(x3)+(0,0.9)$) ; \end{tikzpicture} \par\end{centering}
\par\end{centering}

\protect\caption{Graphical Model Representation of FHMMs.}

\label{fig:fhmm} 
\end{figure}

The Factorial Hidden Markov Model (FHMM) \cite{zoubin} is an extension
of the Hidden Markov Model (HMM), in which the hidden states are factorized
into several independent Markov chains, and emissions (observations)
are determined by their combination. FHMMs have found successful applications
in speech recognition \cite{speech}, source separation \cite{sourcesep,nonfhmm,power,additive},
natural language processing \cite{pos} and bioinformatics \cite{bioinfo}.
A graphical model representation of the FHMM is shown in Fig.\ref{fig:fhmm}. 

Learning FHMMs naturally yields a challenging simultaneous model selection
issue, i.e., how many independent Markov chains we need (layer-level
model selection), and what is the dimensionality (number of hidden
states) of each Markov chain. As the model space increases exponentially
with the number of the Markov chains, it is not feasible to employ
a grid search based method like cross-validation \cite{crossvalid}.
More sophisticated methods, like variational Bayesian inference (VB-FHMMs)
\cite{beal} and the nonparametric Bayesian Infinite FHMMs (iFHMMs)
\cite{ifhmm} have been proposed, but they cannot fully address the
issue. For example, iFHMMs restrict the dimensionality of each Markov
chain to be binary. Further, high computational costs of VB-FHMMs
and iFHMMs restrict their applicability to large scale problems.\vspace{-5bp}

This paper addresses the above model selection issue of FHMMs by extending
Factorized Asymptotic Bayesian (FAB) inference, which is a recent
model selection framework and has shown superior performance than
nonparametric Bayesian inference for Mixture Models \cite{fabmm},
Hidden Markov Models \cite{fabhmm}, Latent Feature Models \cite{fablfm},
and Hierarchical Mixture of Experts Models \cite{fabexpert}. Our
method, namely $\text{FAB}_{fhmm}$, not only fully addresses the
simultaneous model selection issue, but also offers the following
two contributions.\vspace{-5bp}

\textbf{1) Better marginal log-likelihood approximation with partial
marginalization technique}: Previous FAB inference for HMMs \cite{fabhmm}
has applied the Laplace approximations both to emission probabilities
and to transition probabilities. Our approach applies the Laplace
approximations only to emission probabilities, and transition probabilities
are integrated out. As we do not conduct asymptotic approximation
but follow exact marginalization on transition probabilities, we can
better approximate marginal log-likelihood.\vspace{-5bp}

\textbf{2) A quantitative analysis of the shrinkage process}: One
of the strong features in FAB inference is a shrinkage effect, caused
by FAB-unique asymptotic regularization, on hidden variables , i.e.,
redundant hidden states are automatically removed during the FAB EM-like
iterative optimization. Although its strong model selection capability
has been empirically confirmed \noun{\cite{fabmm,fabhmm,fablfm}},
its mathematical behavior has not been well studied. This paper carefully
investigates the shrinkage process, and proves that if there are two
very similar hidden states in FHMMs, then one state would almost surely
``die out''. We reveal the following chain reaction between E-steps
and M-steps: if the variational probabilities of some states are shrunk
in an E-step, then their corresponding model parameters will also
be shrunk in the next M-step, causing these state to be shrunk further
in future E-steps. Moreover, under certain condition, this shrinkage
process is accelerating. This finding also partially answers the parameter
identifiability problem of FHMMs.

\section{Related Work}

\subsection{Factorial HMMs}

The pioneering work of FHMMs was by Z. Ghahramani and M. Jordan \cite{zoubin}.
The idea of FHMMs is to express emissions (observations) by combining
$M$ independent Markov chains. If the individual Markov chain have
binary states, the FHMM can express $2^{M}$ different emission distributions.
Their inference, \emph{structured variational inference}, uses Baum-Welch
algorithm \cite{tutorial} to learn parameters. Generally speaking,
variational Bayesian (VB) inference \cite{grory,beal} can prune redundant
latent states, but previous studies have suggested the pruning effect
is not strong enough to achieve good model selection performance.

Recently, infinite FHMMs (iFHMMs) have been proposed \cite{ifhmm}
to address the model selection issue of FHMMs. iFHMMs employ the Markov
Indian Buffet Process (mIBP) as the prior of the transition matrix
of the infinite hidden states. Although iFHMMs offer strong model
selection ability in learning FHMMs, they have a few limitations.
First, iFHMMs restrict latent variables to be binary while the original
FHMMs \cite{zoubin} have no such limitation. This restriction makes
iFHMMs generate more represented states than variational methods,
tending to overfit the data. In contrast, excluding such a restriction
and allowing different Markov chains to have different numbers of
hidden states may give us better understanding of the data. Second,
the slice sampling used to optimize the iFHMM is considerably slower
than variational methods, and therefore iFHMMs do not scale well to
large scale scenarios.

\subsection{FAB Inference}

Factorized asymptotic Bayesian (FAB) inference has been originally
proposed to address model selection of mixture models \cite{fabmm}.
FAB inference selects the model which maximizes an asymptotic approximation
of the marginal log-likelihood of the observed data, referred to as
\emph{Factorized Information Criterion }(FIC), with an EM-like iterative
optimization procedure. Previous studies have extended FAB inference
to HMMs (sequential) \cite{fabhmm} and Latent Feature Models (factorial)
\cite{fablfm}, and have shown superiority against variational inference
and nonparametric Bayesian methods in terms of model selection accuracy
and computational cost. It is an interesting open challenge to investigate
FAB inference for their intermediate (both sequential and factorial)
models, i.e., FHMMs.

\section{Factorial Hidden Markov Models}

\vspace{-5pt}
Suppose we have observed $N$ independent sequences, denoted as $\boldsymbol{x^{N}}=\boldsymbol{x}^{1},\cdots,\boldsymbol{x}^{N}$.
The n-th sequence is denoted as $\boldsymbol{x}^{n}=\boldsymbol{x}_{1}^{n},\cdots,\boldsymbol{x}_{T_{n}}^{n}$,
where $T_{n}$ is the length of the $n$-th sequence. Respectively,
we denote the corresponding sequences of hidden state variables as
$\boldsymbol{z^{N}}=\boldsymbol{z}^{1},\cdots,\boldsymbol{z}^{N}$,
and each sequence $\boldsymbol{z}^{n}=\boldsymbol{z}_{1}^{n},\boldsymbol{z}_{2}^{n},\cdots,\boldsymbol{z}_{T_{n}}^{n}$.
$\boldsymbol{z}_{t}^{n}$ consists of hidden variables of $M$ independent
HMMs: $\boldsymbol{z}_{t}^{n}=\boldsymbol{z}_{t}^{n,(1)},\boldsymbol{z}_{t}^{n,(2)},\cdots,\boldsymbol{z}_{t}^{n,(M)}$.
The $m$-th HMM has $K_{m}$ hidden states, i.e., $\boldsymbol{z}_{t}^{n,(m)}\in\{0,1\}^{K_{m}}$.
We represent $\boldsymbol{z}_{t}^{n,(m)}$ as a $K_{m}\cdot1$ binary
vector, where the $k$-th component, $z_{t,k}^{n,(m)}$, denotes whether
state $z_{t}^{n,(m)}=k$. 

An FHMM model is specified as the following probability density:\vspace{-8pt}
\begin{align}
p(\boldsymbol{x}^{n},\boldsymbol{z}^{n}|\boldsymbol{\theta})\!=\! & \prod_{m=1}^{M}\!\!\Big\{ p(\boldsymbol{z}_{1}^{n,(m)}|\boldsymbol{\alpha}^{m})\!\prod_{t=2}^{T_{n}}p(\boldsymbol{z}_{t}^{n,(m)}|\boldsymbol{z}_{t-1}^{n,(m)},\boldsymbol{\beta}^{m})\Big\}\nonumber \\
 & \prod_{t=1}^{T}p(\boldsymbol{x}_{t}^{n}|\boldsymbol{z}_{t}^{n},\boldsymbol{\phi}),\label{eq:fhmm}
\end{align}
where $\boldsymbol{\theta}=(\boldsymbol{\alpha},\boldsymbol{\beta},\boldsymbol{\phi})$.
The initial, transition and emission distributions are represented
as $p(\boldsymbol{z}_{1}^{(m)}|\boldsymbol{\alpha}^{m})$, $p(\boldsymbol{z}_{t}^{n,(m)}|\boldsymbol{z}_{t-1}^{n,(m)},\boldsymbol{\beta}^{m})$,
and $p(\boldsymbol{x}_{t}^{n}|\boldsymbol{z}_{t}^{n},\boldsymbol{\phi})$,
respectively.

The $m$-th initial and transition distributions are defined as follows: 
\begin{itemize}
\item $p(\boldsymbol{z}_{1}^{(m)}|\boldsymbol{\alpha}^{(m)})=\prod_{k=1}^{K_{m}}(\alpha_{k}^{(m)})^{z_{1,k}^{(m)}}$; 
\item $p(\boldsymbol{z}_{t}^{(m)}|\boldsymbol{z}_{t-1}^{(m)},\boldsymbol{\beta}^{(m)})=\prod_{j,k=1}^{K_{m}}(\beta_{j,k}^{(m)})^{z_{t-1,j}z_{t,k}}$. 
\end{itemize}
Here $\boldsymbol{\alpha}^{(m)}=(\alpha_{1}^{(m)},\cdots,\alpha_{K_{m}}^{(m)})$,
with $\sum_{k=1}^{K_{m}}\alpha_{k}^{(m)}=1$; $\boldsymbol{\beta}^{(m)}=\left(\beta_{i,j}^{(m)}\right)$,
with each row $\boldsymbol{\beta}_{k}^{(m)}$ summing to 1.

By following the original FHMM work \cite{zoubin}, this paper considers
multidimensional Gaussian emission, while it is not difficult to extend
our discussion to more general distributions like one in the exponential
family. The emission distribution is jointly parameterized across
the hidden states in the $M$ HMMs as follows:

\vspace{-10pt}
\begin{equation}
p(\boldsymbol{x}_{t}^{n}|\boldsymbol{z}_{t}^{n},\phi)=\mathcal{N}(\boldsymbol{x}_{t}^{n},\boldsymbol{\mu}_{t}^{n},\boldsymbol{C}),
\end{equation}
where $\boldsymbol{\mu}_{t}^{n}$ and $\boldsymbol{C}$ are the mean
vector and covariance matrix. A key idea (and difference from HMMs)
in FHMMs is the construction of the mean vector $\boldsymbol{\mu}_{t}^{n}$.
More specifically, the mean vector $\boldsymbol{\mu}_{t}^{n}$ is
represented by the following linear combination: \vspace{-5pt}
\begin{align}
\boldsymbol{\mu}_{t}^{n}= & \sum_{m=1}^{M}\bd W^{m}\boldsymbol{z}_{t}^{n,(m)},
\end{align}
 where $\bd W^{m}$ is a $D\times K_{m}$ matrix. The $k$-th row
of $\bd W^{m}$ is denoted as $\bd W_{k}^{m}$, specifying the contribution
of the $k$-th state in the $m$-th HMM to the mean. Here, $\phi$
in \eqref{eq:fhmm} can be represented as $\phi=(\bd W,\bd C)$.

\section{Refined Factorized Information Criterion for FHMMs}

FIC derivation starts from the following equivalent form of marginal
log-likelihood:\vspace{-10pt}
 \begin{addmargin}{-1em}
\begin{equation}
\log p(\boldsymbol{x^{N}}|\mathcal{M})\equiv\max_{q}\sum_{\boldsymbol{z^{N}}}q(\boldsymbol{z^{N}})\log(\frac{p(\boldsymbol{x^{N}},\boldsymbol{z^{N}}|\mathcal{M})}{q(\boldsymbol{z^{N}})})\label{jensen}
\end{equation}
\end{addmargin}

$\mathcal{M}$ represents a model and $\mathcal{M}=(M,K_{1},\cdots,K_{M})$
in FHMMs. $q(\bd z)$ is a variational distribution over the hidden
states. For a state vector $\bd z{}_{t}^{n,m}$, its expectation under
$q$, $\mathbf{E}_{q}[\bd z_{t}^{n,m}]$, is denoted in shorthand
as $q(\bd z_{t}^{n,m})$, and the expectation of two consecutive states
$\mathbf{E}_{q}[z_{t-1,j}^{n,m}\cdot z_{t,k}^{n,m}]$ as $q(z_{t-1,j}^{n,m},z_{t,k}^{n,m})$.

A direct application of the technique proposed in \cite{fabmm,fabhmm}
leads to the following asymptotic approximation of the complete marginal
log-likelihood: \begingroup \addtolength{\jot}{-0.7em}
\begin{align}
p(\boldsymbol{x^{N}}, & \boldsymbol{z^{N}}|\mathcal{M})=\prod_{m=1}^{M}\Bigl\{\underbrace{\int p(\boldsymbol{z}_{1}^{(m)}|\boldsymbol{\alpha}^{m})p(\bd\alpha^{m}|\mathcal{M})d\bd\alpha^{m}}_{\textnormal{Laplace\ Approx}.}\nonumber \\
 & \underbrace{\int\prod_{t=2}^{T_{n}}p(\boldsymbol{z}_{t}^{n,(m)}|\boldsymbol{z}_{t-1}^{n,(m)},\boldsymbol{\beta}^{m})p(\boldsymbol{\beta}^{m}|\mathcal{M})d\boldsymbol{\beta}^{m}}_{\textnormal{Laplace\ Approx}.}\Bigr\}\nonumber \\
 & \underbrace{\int\prod_{t=1}^{T}p(\boldsymbol{x}_{t}^{n}|\boldsymbol{z}_{t}^{n},\boldsymbol{\phi})p(\boldsymbol{\phi}|\mathcal{M})d\boldsymbol{\phi}}_{\textnormal{Laplace\ Approx}.}.\label{eq:fic}
\end{align}
\endgroup Instead, this paper proposes the FIC derivation with integrating
out the initial and transition distributions. \begingroup \addtolength{\jot}{-0.4em}
\begin{align}
p(\boldsymbol{x^{N}}, & \boldsymbol{z^{N}}|\mathcal{M})=\prod_{m=1}^{M}\Bigl\{\underbrace{\int p(\boldsymbol{z}_{1}^{(m)}|\boldsymbol{\alpha}^{m})p(\bd\alpha^{m}|\mathcal{M})d\bd\alpha^{m}}_{\text{Integrated \ Out}}\nonumber \\
 & \underbrace{\int\prod_{t=2}^{T_{n}}p(\boldsymbol{z}_{t}^{n,(m)}|\boldsymbol{z}_{t-1}^{n,(m)},\boldsymbol{\beta}^{m})p(\boldsymbol{\beta}^{m}|\mathcal{M})d\boldsymbol{\beta}^{m}}_{\text{Integrated \ Out}}\Bigr\}\nonumber \\
 & \underbrace{\int\prod_{t=1}^{T}p(\boldsymbol{x}_{t}^{n}|\boldsymbol{z}_{t}^{n},\boldsymbol{\phi})p(\boldsymbol{\phi}|\mathcal{M})d\boldsymbol{\phi}}_{\text{Laplace \ Approx.}}.\label{eq:rfic}
\end{align}
\endgroup  The original motivation of FIC is to approximate ``intractable''
Bayesian marginal log-likelihood. The key idea in \eqref{eq:rfic}
is to refine approximation by analytically solving the tractable integrations
(w.r.t. $\alpha^{m}$ and $\boldsymbol{\beta}^{m}$) and minimizing
the asymptotic approximation error.

After integrating out all the parameters, and normalizing the Hessians
w.r.t. $\bd W$ and $\boldsymbol{C}^{-1}$, we obtain:

With uninformative conjugate priors on $\bd\alpha^{m}$ and $\boldsymbol{\beta}^{m}$,
we can calculate \eqref{eq:rfic} as follows:

\vspace{-8pt}
\begin{align}
 & p(\bd{x^{N}},\bd{z^{N}}|\mathcal{M})\nonumber \\
\approx & \prod_{m}\Big(\frac{\prod_{k}\Gamma(c_{m,0,k}+1)}{\Gamma(\sum_{k}c_{m,0,k}+K_{m})}\prod_{j}\frac{\prod_{k}\Gamma(c_{m,j,k}+1)}{\Gamma(\sum_{k}c_{m,j,k}+K_{m})}\Big)\nonumber \\
 & \cdot\prod_{d=1}^{D}(2\pi)^{\half{K_{0}+1}}(\prod_{n=1}^{N}\half{T_{n}})^{-\half1}c_{d}^{-\half{K_{0}-2}}|\hat{\mathcal{F}}_{\overbar{\bd W}_{d},\bar{c}_{d}}|^{-\half1}\nonumber \\
 & \cdot\prod_{m,k=1}^{M,K_{m}}(\sum_{n,t=1}^{N,T_{n}}z_{t,k}^{n,(m)})^{-\half1},\label{eq:collapsedPxm}
\end{align}

\vspace{-4pt}
where $K_{0}=\sum_{m}K_{m}$, and $\hat{\mathcal{F}}_{\overbar{\bd W}_{d},\bar{c}_{d}}$
is the normalized Hessian w.r.t. the $d$-th rows of $\bd W$ and
$\boldsymbol{C}^{-1}$ at the maximum likelihood estimators $\overbar{\bd W},\overbar{\bd C}$.
The normalization makes $|\hat{\mathcal{F}}_{\overbar{\bd W}_{d},\bar{c}_{d}}|=O(1)$,
by extracting the terms involving $\bd z$.

\eqref{jensen} computes the expectation of $\log p(\bd{x^{N}},\bd{z^{N}}|\mathcal{M})$
w.r.t. the variational distribution $q$. Taking the logarithm of
\eqref{eq:collapsedPxm}, many log-Gamma terms appear, whose exact
expectations require to enumerate the exponentially many possible
configurations of $\bd z$, which is infeasible. Thus we propose a
first-order approximation to them, based on the following lemmas,
where $\epsilon_{1},\epsilon_{2},\epsilon_{3}$ are small bounded
errors. \begin{lem} Suppose $\{z_{1}^{n},\cdots,z_{T_{n}}^{n}\}_{n=1}^{N}$
are $N$ sequences of Bernoulli random variables. Within each sequence,
$\{z_{t}^{n}\}_{t=1}^{T_{n}}$ are independent with each other. Let
$y_{n}=\sum_{i}z_{i}^{n}$, $\bar{y}_{n}=E[y_{n}]$. Besides, there
are $N$ numbers $\{\hat{y}_{n}\}$, $\forall k,\hat{y}_{n}\approx\bar{y}_{n}$.
When all $T_{n}$ are large enough:

1) $\mathbf{E}[\log\Gamma(y_{n})]=y_{n}\log\bar{y}_{n}-(\bar{y}_{n}+\half1\log\bar{y}_{n})+\log2\pi+\epsilon_{1}.$

2) $\mathbf{E}\left[\ssum_{n}\!\log\!\Gamma(y_{n})\!-\!\log\!\Gamma(\ssum_{n}y_{n})\right]\!=\!\ssum_{n}y_{n}\!\log\!\big(\frac{\hat{y}_{n}}{\ssum_{m=1}^{N}\hat{y}_{m}}\big)\allowbreak+\half1\log\big(\ssum_{n}\hat{y}_{n}\big)-\half1\ssum_{n}\log\hat{y}_{n}+(N-1)\log2\pi+\epsilon_{2}.$

3) $\mathbf{E}[\log y_{n}]=\log\hat{y}_{n}+\frac{1}{\hat{y}}(\bar{y}_{n}-\hat{y}_{n})+\epsilon_{3}.$
\end{lem} Proof can be found in the Appendix.

As defined above, $c_{m,j,k}=\sum_{n,t=1}^{N,T_{n}-1}z_{t,j}^{n,m}z_{t+1,k}^{n,m}$.
Since $\{z_{t}^{n,m}\}_{t=1}^{T_{n}}$ follows a Markov process, the
dependency between $z_{t,j}^{n,m}$ and $z_{t+\Delta t,j}^{n,m}$
vanishes quickly when the time gap $\Delta t$ increases, leading
to the uncoupling between $z_{t,j}^{n,m}z_{t+1,k}^{n,m}$ and $z_{t+\Delta t,j}^{n,m}z_{t+1+\Delta t,k}^{n,m}$.
Thus we can approximately regard $c_{m,j,k}$ as the sum of independent
Bernoulli random variables, and therefore Lemma 1 apply. The same
argument applies to $c_{m,0,k}$.

In order to avoid recursion relations w.r.t. $q$ during the inference,
we introduce an auxiliary distribution $\hat{q}$, which is always
close to $q$, in that $\hat{q}(z_{t,k}^{n,m})\approx q(z_{t,k}^{n,m}),\hat{q}(z_{t,j}^{n,m},z_{t+1,k}^{n,m})\approx q(z_{t,j}^{n,m},z_{t+1,k}^{n,m})$.
The Gamma terms in \eqref{eq:collapsedPxm} will be expanded about
the expectations of their parameters w.r.t. $\hat{q}$. We will discuss
how to choose $\hat{q}$ in Section \ref{sub:FABVEM}.

Taking the logarithm of \eqref{eq:collapsedPxm}, applying Lemma 1,
and dropping asymptotically small terms (including the determinants
of the normalized Hessian matrices), we derive the approximation of
$p(\bd{x^{N}},\bd{z^{N}}|\mathcal{M})$. Plugging it into \eqref{jensen},
we obtain an asymptotic approximation of the lower bound of $\log p(\boldsymbol{x^{N}}|M)$:

\hspace*{-4pt}\vbox{
\begin{align}
 & \log p(\bd{x^{N}}|\mathcal{M})\mathcal{\ge J}(q,\boldsymbol{x^{N}})\nonumber \\
\approx & \sum q(\bd{z^{N}})\bigg[\sum_{n,t}\Big(\log p(\bd x_{t}^{n}|\bd z_{t}^{n},\overbar{\bd W},\overbar{\bd C})+\sum_{m,k}z_{t,k}^{n,m}\log\delta_{k}^{m}\Big)\nonumber \\
 & +\sum_{m,n=1}^{M,N}\Big(\sum_{k}z_{0,k}^{n,m}\log\hat{\alpha}_{k}^{m}+\sum_{t=1}^{T_{n}-1}\sum_{j,k}z_{t,j}^{n,m}z_{t+1,k}^{n,m}\log\hat{\beta}_{j,k}^{m}\Big)\nonumber \\
 & -\log q(\bd{z^{N}})\bigg]\underline{+\sum_{m,n,t}\log\Delta^{m}-\half D\sum_{m,k=1}^{M,K_{m}}\log(\hat{c}_{m,\cdot,k})}\nonumber \\
 & \underline{-\half1\sum_{m,k}\log(\hat{c}_{m,0,k}+1)-\half1\sum_{m,j,k}\log(\hat{c}_{m,j,k}+1)}\nonumber \\
 & \underline{+\half1\sum_{m,j}\log(\sum_{k}\hat{c}_{m,j,k}+K_{m})+\half1\sum_{m}\log(N+K_{m})}\nonumber \\
 & \underline{+\frac{D}{2}\sum_{m}K_{m}+\sum_{m}(K_{m}^{2}-1)\log2\pi+\epsilon,}\label{collapsedFICBound}
\end{align}

}

with the definitions 
\begin{align}
\hat{q}(z_{t,i}^{n,m}) & =E_{q}(z_{t,i}^{n,m}),\quad q(z_{t,j}^{n,m},z_{t+1,k}^{n,m})=E_{q}(z_{t,j}^{n,m},z_{t+1,k}^{n,m}),\nonumber \\
\hat{c}_{m,0,k} & =\sum_{n=1}^{N}\hat{q}(z_{1,k}^{n,m}),\quad\hat{c}_{m,\cdot,k}=\sum_{n,t=1}^{N,T_{n}}\hat{q}(z_{t,k}^{n,m}),\nonumber \\
\hat{c}_{m,j,k} & =\sum_{n,t=1}^{N,T_{n}-1}\hat{q}(z_{t,j}^{n,m},z_{t+1,k}^{n,m}),\allowdisplaybreaks\nonumber \\
\delta_{k}^{m} & =\frac{1}{\Delta^{m}}\exp\{-\frac{D}{2\sum_{n,t=1}^{N,T_{n}}\hat{q}(z_{t,k}^{n,m})}\},\nonumber \\
\hat{\alpha}_{k}^{m} & =\frac{1+\sum_{n=1}^{N}\hat{q}(z_{1,k}^{n,m})}{K_{m}+N},\nonumber \\
\hat{\beta}_{j,k}^{m} & =\frac{1+\sum_{n,t=1}^{N,T_{n}-1}\hat{q}(z_{t,j}^{n,m},z_{t+1,k}^{n,m})}{K_{m}+\sum_{n,t=1}^{N,T_{n}-1}q(z_{t,j}^{n,m})},\label{eq:collapsedDelta}
\end{align}

and $\Delta^{m}$ is a normalization constant that makes $\sum_{k=1}^{K_{m}}\delta_{k}^{m}=1$,
$\varepsilon$ is a small constant error bound of the approximations.
Note the ML estimators $\overbar{\bd W},\overbar{\bd C}$ inside the
summation are subject to the specific instantiation of $\bd{z^{N}}$.

In \eqref{eq:collapsedDelta}, $\hat{\alpha}$ and $\hat{\beta}$
can be viewed as the estimated initial transition probabilities of
the FHMM w.r.t. $\hat{q}$.

The FIC of the FHMM, denoted by $\text{FIC}_{fhmm}$, is obtained
by maximizing $\mathcal{J}(q,x^{N})$ w.r.t. $q$: 
\begin{equation}
\text{FIC}(\boldsymbol{x^{N}},\mathcal{M})=\max_{q}\{\mathcal{J}(q,\boldsymbol{x^{N}})\}.
\end{equation}

Similar to $\text{FIC}_{mm}$ and $\text{FIC}_{hmm}$, the use of
$\text{FIC}_{fhmm}$ as the approximation of the observed data log-likelihood
under a certain model is justified: \begin{thm} $\text{FIC}(\boldsymbol{x^{N}},\mathcal{M})$
is asymptotically equivalent to $\log p(\boldsymbol{x^{N}}|\mathcal{M})$.
\end{thm} The proof is analogous to that of $\text{FIC}_{mm}$ and
$\text{FIC}_{hmm}$\cite{fabmm,fabhmm}, and omitted here.

\section{FAB for FHMMs}

\subsection{FAB's Lower Bound of FIC}

Since $\overbar{\bd W},\overbar{\bd C}$ depends on the specific instantiation
of $\boldsymbol{z^{N}}$, in order to compute $\text{FIC}_{fhmm}$
exactly, we need to evaluate $\overbar{\bd W},\overbar{\bd C}$ for
each instantiation of $\boldsymbol{z^{N}}$, which is infeasible.
So we bound $\text{FIC}(\boldsymbol{x^{N}},M)$ from below by setting
all $\overbar{\bd W},\overbar{\bd C}$ to the same values, and get
a relaxed lower bound $\mathcal{G}$:

$\mathcal{G}(q,\boldsymbol{x^{N}},\boldsymbol{W},\bd C)<\mathcal{J}(q,\boldsymbol{x^{N}})\le\log p(\bd{x^{N}}|\mathcal{M}).$

$\text{FAB}_{fhmm}$ is an optimization procedure that tries to maximize
the above lower bound of $\text{FIC}_{fhmm}$: 
\begin{equation}
\mathcal{M}^{*},q^{*},\bd W^{*},\bd C^{*},=\argmax_{\mathcal{M},q,\bd W,\bd C}\mathcal{G}(q,\boldsymbol{x^{N}},\boldsymbol{W},\bd C).\label{fabobj}
\end{equation}

\subsection{FAB Variational EM Algorithm\label{sub:FABVEM}}

Let us first fix the structural parameters of the FHMM model, i.e.
$M,K_{1},\cdots,K_{M}$. Then our objective is to optimize \eqref{fabobj}
w.r.t. $(q,\boldsymbol{W},\bd C)$. In this phase FAB is a typical
variational EM Algorithm, iterating between E-steps and M-steps. We
denote the $i$-th iteration with the superscript $\{i\}$.

In \eqref{collapsedFICBound}, the auxiliary distribution $\hat{q}$
is required to be close to $q$. So during the FAB optimization, we
always set $\hat{q}$ to be the variational distribution in the previous
iteration, i.e. $\hat{q}^{\{i\}}=q^{\{i-1\}}$.

In the $i$-th E-step, we fix $\boldsymbol{W},\bd C=\boldsymbol{W}^{\{i-1\}},\bd C^{\{i-1\}}$,
and $\hat{q}=q^{\{i-1\}}$. The E-step computes the $q$ which maximizes
\eqref{fabobj}.

The exact E-step for FHMMs is intractable \cite{zoubin}, and thus
in E-step, we adopt the \emph{Mean-field Variational Inference} proposed
in \cite{zoubin}. 

\subsubsection{FAB E-Step}


\paragraph{Mean-Field Variational Inference}

The \emph{Mean-Field Variational Inference }approximates the FHMM
with $M$ uncoupled HMMs, by introducing a variational parameter $\boldsymbol{h}_{t}^{n,(m)}$
for each layer variable $\boldsymbol{z}_{t}^{n,(m)}$. $\boldsymbol{h}_{t}^{n,(m)}$
approximates the contribution of $\boldsymbol{z}_{t}^{n,(m)}$ to
the corresponding observation $\boldsymbol{x}_{t}^{n}$.

The structured variational distribution $q$ is defined as 
\begin{equation}
q(\boldsymbol{z^{N}}\!|\boldsymbol{h})\!=\!\frac{1}{\mathcal{Z}}\!\!\prod_{n,m=1}^{N,M}\!\!\!\!\Big\{\hat{q}(\boldsymbol{z}_{1}^{n,(m)}|\boldsymbol{h})\!\prod_{t=2}^{T_{n}}\hat{q}(\boldsymbol{z}_{t}^{n,(m)}|\boldsymbol{z}_{t-1}^{n,(m)},\boldsymbol{h})\!\Big\},\label{vardist}
\end{equation}
where $\mathcal{Z}$ is the normalization constant, and 
\begin{align}
\hat{q}(\boldsymbol{z}_{1}^{n,(m)}|\boldsymbol{h}) & =\prod_{k=1}^{K_{m}}(h_{1,k}^{n,(m)}\hat{\alpha}_{k}^{n,(m)})^{z_{1,k}^{n,(m)}},\nonumber \\
\hat{q}(\boldsymbol{z}_{t}^{n,(m)}|\boldsymbol{z}_{t-1}^{n,(m)},\boldsymbol{h}) & =\prod_{k=1}^{K_{m}}\Big(h_{t,k}^{n,(m)}\prod_{j=1}^{K_{m}}(\hat{\beta}_{j,k}^{(m)})^{z_{t-1,j}^{n,(m)}}\Big)^{z_{t,k}^{n,(m)}}.\label{vardist3}
\end{align}
Note $\boldsymbol{h}_{t}^{n,(m)}$ is a $1\times K_{m}$ vector, which
gives a bias for each of the $K_{m}$ settings of $z_{t,k}^{n,(m)}$.

We obtain a system of equations for $\boldsymbol{h}_{t}^{n,(m)}$
and $\boldsymbol{q}(z_{t}^{n,(m)})$, which minimize $\text{KL}(q||\hat{p}(\boldsymbol{z^{N}}|\boldsymbol{x^{N}},\boldsymbol{\theta}))$,
by setting its derivation w.r.t. $\boldsymbol{h}_{t}^{n,(m)}$ to
0: 
\begin{equation}
\boldsymbol{h}_{t}^{n,(m)}=\textnormal{diag}\{\boldsymbol{\delta}^{m}\}\exp\Big\{\bd W^{m'}\boldsymbol{C}^{-1}\tilde{\boldsymbol{x}}_{t}^{n,(m)}-\frac{1}{2}\Lambda^{m}\Big\},\label{updateH}
\end{equation}
where $\Lambda^{m}$ is the vector consisting of the diagonal elements
of $\bd W^{m'}\boldsymbol{C}^{-1}\bd W^{m}$; $\textnormal{diag}(\boldsymbol{v})$
is an operator that constructs a matrix whose diagonal is $\boldsymbol{v}$,
and all off-diagonal elements are 0; $\tilde{\boldsymbol{x}}_{t}^{n,(m)}$
is the residual:\vspace{-8pt}
 
\begin{equation}
\tilde{\boldsymbol{x}}_{t}^{n,(m)}=\boldsymbol{x}_{t}^{n}-\sum_{l\ne m}^{M}W^{l}q(\bd z_{t}^{n,(l)}).\label{tildeX}
\end{equation}
In \eqref{updateH}, $\boldsymbol{h}$ depends on $q$, but $q$ also
depends on $\boldsymbol{h}$ in the Forward-Backward routine, in a
complicated way. Such an interdependence makes the exact solution
difficult to find. Therefore in \eqref{tildeX}, we use $q^{\{i-1\}}$
to get approximate solutions of $\bd h$.

Note the bias $h_{t,k}^{n,(m)}$ incorporates the effects of the shrinkage
factor $\delta_{k}^{m}$. Through the bias, the shrinkage factors
will induce the \emph{Hidden State Shrinkage} to be discussed later.

\paragraph{Forward-Backward Routine}

After updating the variational parameters $\boldsymbol{h}$, we use
the Forward-Backward algorithm to compute the sufficient statistics
$q(\boldsymbol{z}_{t}^{n,(i)})$, and $q(\boldsymbol{z}_{t-1}^{n,(i_{1})}\boldsymbol{z}_{t}^{n,(i_{2})'})$.

The following recurrence relations for the forward quantities $f_{t,k}^{n,(m)}$
and backward quantities $b_{t,k}^{n,(m)}$ is derived from the definition
(\ref{vardist}-\ref{vardist3}) of $q$: \begin{addmargin}{-1em}
\begin{align}
f_{t,k}^{n,(m)} & =\begin{cases}
\frac{1}{\zeta_{1}^{n,(m)}}h_{1,k}^{n,(m)}\hat{\alpha}_{k}^{(m)} & \mathsmaller{\textnormal{if }t=1}\\
\frac{1}{\zeta_{t}^{n,(m)}}h_{t,k}^{n,(m)}\ssum_{j=1}^{K_{m}}f_{t-1,j}^{n,(m)}\hat{\beta}_{j,k}^{(m)} & \mathsmaller{\textnormal{if }t>1}
\end{cases}\nonumber \\
b_{t,k}^{n,(m)} & =\begin{cases}
\frac{1}{\zeta_{t+1}^{n,(m)}}\ssum_{j=1}^{K_{m}}\hat{\beta}_{k,j}^{(m)}h_{t+1,j}^{n,(m)}b_{t+1,j}^{n,(m)} & \mathsmaller{\textnormal{if }t<T_{n}}\\
1 & \mathsmaller{\textnormal{if }t=T_{n}}
\end{cases}\label{eq:fbrecurrence}
\end{align}
\end{addmargin}where $\zeta_{t}^{n,(m)}$ is the normalization constant
that makes $\sum_{k=1}^{K_{m}}f_{t,k}^{n,(m)}=1$.

The sufficient statistics are computed thereafter: 
\begin{align}
q(z_{t,k}^{n,(m)}) & =f_{t,k}^{n,(m)}b_{t,k}^{n,(m)},\label{statqZ}\\
q(z_{t-1,j}^{n,(m)},\, z_{t,k}^{n,(m)}) & =\frac{1}{\zeta_{t}^{n,(m)}}f_{t-1,j}^{n,(m)}\,\hat{\beta}_{j,k}^{(m)}h_{t,k}^{n,(m)}b_{t,k}^{(m)}.\label{statqZZ}
\end{align}

\subsubsection{M-Step}

In the M-step, we fix $q=\hat{q}=q^{\{i\}}$, and maximize \eqref{fabobj}
w.r.t. $\boldsymbol{W},\boldsymbol{C}$, obtaining their optimal values:

\hspace*{-8pt}\vbox{ 
\begin{align}
 & \boldsymbol{C}\!=\!\textnormal{mdiag}\Big\{\frac{1}{{\displaystyle \sum_{n}\! T_{n}}}\sum_{n,t=1}^{N,T_{n}}\!\!\Big(\bd x_{t}^{n}\bd x_{t}^{n'}\!\!-\!\!\sum_{m=1}^{M}W^{m}q(\boldsymbol{\bd z}_{t}^{n,(m)})\boldsymbol{x}_{t}^{n'}\Big)\!\!\Big\}\nonumber \\
 & \boldsymbol{W}=\left(\sum_{n,t=1}^{N,T_{n}}\boldsymbol{x}_{t}^{n}q(\boldsymbol{z}_{t}^{n'})\right)\left(\sum_{n,t=1}^{N,T_{n}}\boldsymbol{E}_{q}\Big[\boldsymbol{z}_{t}^{n}\boldsymbol{z}_{t}^{n'}\Big]\right)^{\dagger},
\end{align}
} where: 1) In the equation of $\boldsymbol{W}$, $\dagger$ is the
Moore-Penrose pseudo-inverse, $\boldsymbol{W}=\left(\bd W^{1}\cdots\bd W^{M}\right)$,
$\boldsymbol{z}_{t}^{n}$ is an $(\sum_{m=1}^{M}K_{m})\times1$ vector
of the concatenation of $\boldsymbol{z}_{t}^{n,(1)},\cdots,\boldsymbol{z}_{t}^{n,(M)}$;
2) In the equation of $\boldsymbol{C}$, the $\textnormal{mdiag}(X)$
operator sets all the off-diagonal elements of matrix $X$ to 0, leaving
the diagonal ones intact.

\subsection{Automatic Hidden State Shrinkage}

The shrinkage factors $\delta_{k}^{m}$ induce the \emph{hidden state
shrinkage} effect, which is a central mechanism to $\text{FAB}_{fhmm}$'s
model selection ability. Previous FAB methods \cite{fabmm,fabhmm,fablfm}
report similar effects.

In the beginning, we initialize the model with a large enough number
$M$ of layers, and all layers with a large enough number $K_{\text{max}}$
of hidden state values. During the EM iterations, suppose there are
two similar hidden states $i,j$ in the $m$-th layer, and state $i$
is less favored than state $j$, i.e. $\sum_{t}q(z_{t,i}^{(m)})<\sum_{t}q(z_{t,j}^{(m)}).$
According to \eqref{eq:collapsedDelta}, $\delta_{i}^{m}<\delta_{j}^{m}$.
In the next E-step, the variational parameters of the $i$-th state
$\{h_{t,i}^{(m)}\}_{t}$ are more down-weighted than those of the
$j$-th state $\{h_{t,j}^{(m)}\}_{t}$, according to \eqref{updateH}.
From the Forward-Backward update equations \eqref{eq:fbrecurrence}
one can see $q(z_{i}^{(m)})$ decreases with $h_{t,i}^{(m)}$. Therefore
$\{q(z_{i}^{(m)})\}$ decreases more than $\{q(z_{j}^{(m)})\}$, which
in turn causes smaller transition probabilities into state $i$ in
the next M-step. Subsequently, smaller transition probabilities into
state $i$ make $\sum_{t}q(z_{t,i}^{(m)})$ and $\delta_{i}^{m}$
even smaller compared to $\sum_{t}q(z_{t,j}^{(m)})$ and $\delta_{k}^{m}$,
respectively. This process repeats and reinforces itself, making state
$i$ less and less important. When $\sum_{t}q(z_{t,i}^{(m)})<L$,
a pre-specified threshold, then $\text{FAB}_{fhmm}$ regards state
$i$ as redundant and removes it. If all but one states of a layer
are removed, then we know this layer is redundant and $\text{FAB}_{fhmm}$
removes this layer. When $\text{FAB}_{fhmm}$ converges, we obtain
a more parsimonious model with the number of layers $M'$, the number
of hidden states in each layers $K_{1}^{*},\cdots,K_{M'}^{*}$, the
optimal model parameters $\bd W^{*},\bd C^{*}$, and the variational
distribution $q^{*}$. 

This intuition is formulated in the following theorem.

{} \begin{thm} Suppose we have one sufficiently long training sequence
of observations $\bd x_{\bd T}=\bd x_{1},\cdots,\bd x_{T},T\gg1$.
In the $m$-th layer, two states $i,j$ are initialized with proportional
initial and inwards transition probabilities, i.e., $\hat{\alpha}_{i}=\rho_{0}\hat{\alpha}_{j},$
and $\forall k,\hat{\boldsymbol{\beta}}_{k,i}=\rho_{0}\hat{\boldsymbol{\beta}}_{k,j},\rho<1$,
and identical outwards transition probabilities $\forall k,\hat{\boldsymbol{\beta}}_{i,k}=\hat{\boldsymbol{\beta}}_{j,k}$.
Then: 1) after the first EM iteration, the two states have identical
weight vectors and outwards probabilities: $W_{i}^{m}=W_{j}^{m}$,$\forall k,\hat{\boldsymbol{\beta}}'_{i,k}=\hat{\boldsymbol{\beta}}'_{j,k}$,
and proportional inwards probabilities: $\hat{\alpha}'_{i}=\rho_{1}\hat{\alpha}'_{j},$
$\forall k,\hat{\boldsymbol{\beta}}'_{k,i}=\rho_{1}\hat{\boldsymbol{\beta}}'_{k,j}$,
but $\rho_{1}<\rho_{0}$. This trend preserves in all the following
iterations. 2) When the probabilities of state $j$ stablize, this
shrinkage process on state $i$ is accelerating: after the $n$-th
iteration, the probability ratio $\rho_{n}$ satisfy: $\rho_{n}/\rho_{n-1}<\rho_{n-1}/\rho_{n-2}$.
\end{thm} \textbf{Proof}. For notation similicity, we drop the superscript
$n$ for the sequence number.

1) For initialization all variational parameters $\bd h$ are set
to $1$. From the Forward-Backward recurrence relations \eqref{eq:fbrecurrence},
it is easily seen that $f_{t,i}^{(m)}=\rho_{0}f_{t,j}^{(m)},b_{t,i}^{(m)}=b_{t,j}^{(m)}$.
Thus from \eqref{statqZ},\eqref{statqZZ}, the sufficient statistics
of the variational probability $q$ satisfy, for $\forall t$: \vspace{-10pt}
\begin{addmargin}{-1em} 
\begin{align}
q(z_{t,i}^{(m)})\! & =\!\rho_{0}q(z_{t,j}^{(m)}),\, q(z_{t-1,i}^{(m)}\, z_{t,k}^{(m)})\!=\!\rho_{0}q(z_{t-1,j}^{(m)}\, z_{t,k}^{(m)})\!\!\label{eq:ratioqij}
\end{align}
\end{addmargin}\vspace{-5pt}
In the M-step, when updating $\bd W$, one relevant equation is

\vspace{-10pt}
\begin{addmargin}{-1em} 
\begin{align}
\ssum_{t}\bd x_{t}q(\bd z_{t}^{m})'\!=\! & \!\ssum_{l\ne m}\bd W^{l}\!\ssum_{t}q(\bd z_{t}^{l})q(\bd z_{t}^{m})'\!+\!\bd W^{m}\!\ssum_{t}\textnormal{diag}\{q(\bd z_{t}^{m})\}\label{eq:updateWm}
\end{align}
\end{addmargin}\vspace{-5pt}

We single out the $i$-th and $j$-th columns of both sides of \eqref{eq:updateWm}:

\vspace{-10pt}
\begin{addmargin}{-1em} 
\begin{align}
\ssum_{t}q(z_{t,i}^{m})\bd x_{t} & \!=\!\!\ssum_{l\ne m}\bd W^{l}\!\ssum_{t}q(z_{t,i}^{m})q(\bd z_{t}^{l})\!+\!\textnormal{\ensuremath{\ssum_{t}}diag}\{q(z_{t,i}^{m})\}\bd W_{i}^{m}\nonumber \\
\ssum_{t}q(z_{t,j}^{m})\bd x_{t} & \!=\!\!\ssum_{l\ne m}\bd W^{l}\!\ssum_{t}q(z_{t,j}^{m})q(\bd z_{t}^{l})\!+\!\textnormal{\ensuremath{\ssum_{t}}diag}\{q(z_{t,j}^{m})\}\bd W_{j}^{m}\label{eq:qw_ij}
\end{align}
\end{addmargin}

By plugging \eqref{eq:ratioqij} into \eqref{eq:qw_ij}, we find $\bd W_{i}^{m}=\bd W_{j}^{m}$.

In the next E-step, using \eqref{updateH}, we can identify 
\begin{align}
\frac{h_{t,i}^{n,(m)}}{h_{t,j}^{n,(m)}}=\frac{\delta_{i}^{m}}{\delta_{j}^{m}} & =\exp\{\frac{D}{2\sum_{t}q(z_{t,j}^{n,m})}-\frac{D}{2\sum_{t}q(z_{t,i}^{n,m})}\}\nonumber \\
 & =\exp\{-\big(\frac{1-\rho_{0}}{\rho_{0}}\big)\frac{D}{2\sum_{t}q(z_{t,j}^{n,m})}\},\label{eq:hratio}
\end{align}

which is constant for any $t$. We denote it as $\lambda_{0}$. Since
$\rho_{0}<1$, obviously $\lambda_{0}<1$.

In the Forward-Backward routine, by examining equations \eqref{eq:fbrecurrence},
one can see $f_{t,i}^{(m)}=\rho_{0}\lambda_{0}f_{t,j}^{(m)},b_{t,i}^{(m)}=b_{t,j}^{(m)}$.
Let $\rho_{1}=\rho_{0}\lambda_{0}$. Again from \eqref{statqZ},\eqref{statqZZ},
the sufficient statistics of the new variational distribution $q'$
satisfy, for $\forall t$: 
\begin{align*}
q'(z_{t,i}^{(m)}) & =\rho_{1}q'(z_{t,j}^{(m)}),\\
q'(z_{t-1,i}^{(m)},z_{t,k}^{(m)}) & =\rho_{1}q'(z_{t-1,j}^{(m)},z_{t,k}^{(m)}),\\
q'(z_{t,k}^{(m)},z_{t+1,i}^{(m)}) & =\rho_{1}q'(z_{t,k}^{(m)},z_{t+1,j}^{(m)}).
\end{align*}

When we update $\hat{\alpha}$ and $\hat{\beta}$, since the training
sequence is sufficiently long ($T\gg1$), the pseudocount ``$1$''
and ``$K_{m}$'' in their update equations can be ignored. Therefore
one can easily obtain
\begin{align*}
\hat{\alpha}'_{i} & =\rho_{1}\hat{\alpha}'_{j},\\
\forall k,\hat{\boldsymbol{\beta}}'_{k,i} & =\rho_{1}\hat{\boldsymbol{\beta}}'_{k,j},\\
\forall k,\hat{\boldsymbol{\beta}}'_{i,k} & =\hat{\boldsymbol{\beta}}'_{j,k}.
\end{align*}

Obviously $\rho_{1}<\rho_{0}$. That is, the new ratio of the initial/inwards
probabilities becomes smaller, while the outwards probabilities remain
unchanged.

The same argument holds for all subsequent iterations.

2) In the $n$-th iteration, let the ratio in \eqref{eq:hratio} be
denoted as $\lambda_{n}$. From the above arguments, $\rho_{n}/\rho_{n-1}=\lambda_{n}$,
thus it is sufficient to prove $\lambda_{n}<\lambda_{n-1}$.

If state $j$ ``survives'' the shrinkage, then after many iterations
its probabilities would become stablized, i.e., $\sum_{t}q(z_{t,j}^{n,m})$
is almost the same between the $n$-th and ($n$-$1$)-th iterations.
In this condition, $\lambda_{n}$ decreases with $\rho_{n}$, together
with $\rho_{n}<\rho_{n-1}$, one can conclude $\lambda_{n}<\lambda_{n-1}$.
$\qedsymbol$

Theorem 5 sheds light on $\text{FAB}_{fhmm}$'s shrinkage mechanism:
given two similar components, if they are not initialized evenly (which
is almost always the case due to randomness), then they will ``compete''
for probability mass, until one dominates the other. Another important
observation is, the M-step ``relays'' the shrinkage effect between
consecutive E-steps, and thus running multiple E-steps in a row would
not gain much speed-up as observed on Latent Feature Models \cite{fablfm}.

\subsection{Parameter Identifiability}

Although HMMs/FHMMs are identifiable in the level of equivalence classes
\cite{mlhmm}, generally they are not identifiable if two states have
very similar outwards transition probabilities and emission distributions.
Formally, if states $i,j$ satisfy: $\forall k,\boldsymbol{\beta}_{i,k}\approx\boldsymbol{\beta}_{j,k}$,
and $p(x|z=i)\approx p(x|z=j)$ , then we can alter the inwards transition
probabilities of $i,j$ without changing the probability law of this
model, as long as the sum of their inwards transition probabilities
is fixed, i.e., $\boldsymbol{\hat{\beta}}_{\cdot,i}+\hat{\boldsymbol{\beta}}_{\cdot,j}=\boldsymbol{\beta}_{\cdot,i}+\boldsymbol{\beta}_{\cdot,j}.$
This forms an equivalence class with infinitely many parameters settings
(one extreme is state $i$ is absorbed into $j$), all of which fit
the observed data equally well. Thus traditional point estimation
methods are unable to compare these parameter settings and pick better
ones.

In contrast, Theorem 5 reveals that, $\text{FAB}_{fhmm}$ will almost
surely shrink and remove one of such two states, picking a parsimonious
parameter setting (e.g. the one with $i$ absorbed into $j$) among
this equivalence class.

\section{Synthetic Experiments}

To test our inference algorithm, we ran experiments on a synthetic
data set. The performance of $\text{FAB}_{fhmm}$ (denoted as ``RFAB'',
i.e. \emph{Refined }FAB) and three major competitors, i.e., conventional
FAB without marginalization (denoted as ``FAB''), variational Bayesian
FHMM (denoted as ``VB''), and iFHMM, was compared.

\subsection{Experimental Settings}

We constructed a ground truth model FHMM, which has 3 layers, with
state numbers $(2,2,3)$. These states comprise a total state space
of $2\cdot2\cdot3=12$ combinations.The observations are 3-dimensional,
and the covariance matrix $\Sigma=\left(\begin{smallmatrix}0.4 & 0 & 0\\
0 & 0.4 & 0\\
0 & 0 & 0.4
\end{smallmatrix}\right)$. The three mean matrices $\{\bd W^{m}\}$, initial probabilities
$\{\bd\alpha^{m}\}$ and transition matrices $\{\bd\beta^{m}\}$ were
randomly generated, and omitted here.

Two sequences of length $T=2000$ were randomly generated. One sequence
was used for training, and the other was used for testing.

RFAB, FAB and VB-FHMM are all initialized with 3 HMMs, and 10 states
in each HMM. Compared to the true model, this setting has a lot of
redundant states. For VB-FHMM, we used the ``component death'' idea
in \cite{beal}, i.e. if a hidden state receives too little probability
mass, then it will be removed. This scheme is similar to FAB's pruning
scheme, despite the fact that the inference algorithm does not incorporate
the shrinkage regularization.

We obtained the Matlab code of iFHMM from Van Gael. We used its default
hyperparameters, i.e.: $\alpha\sim\Gamma(1+K,1+H_{T})$, where $K$
is the number of hidden states, and $H_{T}$ is the $T$-th harmonic
number; $\sigma_{Y},\sigma_{W}\sim\Gamma(1,1)$. In order to see how
iFHMM performs the model selection on different hidden state spaces,
we initialized iFHMM with hidden state numbers $K\in\{4,7,10,15\}$. 

The maximal iterations of all models other than iFHMM are set to 1000.
iFHMM is set to have 500 iterations for burn-in and 500 iterations
for sampling. Each algorithm was tested for 10 trials.

\subsection{Results and Discussions}

Three indices of the models were collected: 1) the eventually obtained
state numbers; 2) the log-likelihood of the training sequence; and
3) the predictive log-likelihood of the test sequence. 

The state number in the three HMMs are first sorted from smallest
to largest, and then averaged. The average values, and their standard
deviations (in parentheses) are reported in the following table:

\begin{table}

\vspace{5pt}

\begin{centering}
\begin{tabular}{|c|c|c|c|}
\multicolumn{1}{c}{} & \multicolumn{1}{c}{$K_{1}$} & \multicolumn{1}{c}{$K_{2}$} & \multicolumn{1}{c}{$K_{3}$}\tabularnewline
\hline 
\hline 
RFAB & 1.9 (0.4) & 3.0 (0.6) & 4.8 (1.1)\tabularnewline
\hline 
FAB & 1.4 (0.6) & 2.2 (0.7) & 4.0 (1.5)\tabularnewline
\hline 
VB & 3.2 (1.5) & 4.7 (1.2) & 6.3 (1.8)\tabularnewline
\hline 
iFHMM/4 & 4.4 (0.3) & / & /\tabularnewline
\hline 
iFHMM/7 & 7.3 (0.7) & / & /\tabularnewline
\hline 
iFHMM/10 & 9.5 (1.2) & / & /\tabularnewline
\hline 
iFHMM/15 & 13.6 (1.3) & / & /\tabularnewline
\hline 
\end{tabular}
\par\end{centering}

\protect\caption{Average State Numbers}
\end{table}

From Table 1, we can see that FAB obtained the most parsimonious learned
models, but often it removes too many states, such that one HMM has
only one state left. In contrast, RFAB is more ``conservative'',
as it keeps more states. This difference is probably because after
marginalizing $\bd\alpha$ and $\bd\beta$, the pseudocount ``1''
appears in the update equations \ref{eq:collapsedDelta} of $\hat{\bd\alpha}$
and $\hat{\bd\beta}$ , which smooths their estimated values. VB-FHMM
removed some states, but there are still a few redundant states left.
It indicates that without the shrinkage regularization, the model
selection ability is limited. iFHMM almost always stays around its
initial number of hidden states, no matter how many states they are
initialized with. This indicates the model selection ability of iFHMM
is weak. 

The training and testing data log-likelihoods are compared in Table
2. RFAB achieves the best log-likelihoods on both sequences, followed
by FAB. VB-FHMM achieves better log-likelihood than FAB on the training
sequence, but worse on the test sequence, which suggests that it overfits
the training data. iFHMM's performance is significantly inferior in
general, and degrades quickly as the initial $K$ increases.

\begin{table}

\vspace{5pt}

\begin{centering}
\begin{tabular}{|c|c|c|}
\multicolumn{1}{c}{} & \multicolumn{1}{c}{Training} & \multicolumn{1}{c}{Testing}\tabularnewline
\hline 
\hline 
RFAB & -\textbf{10937} (254) & \textbf{-12416} (425)\tabularnewline
\hline 
FAB & -11848 (322) & -13647 (598)\tabularnewline
\hline 
VB & -11239 (336) & -14430 (662)\tabularnewline
\hline 
iFHMM/4 & -12256 (293) & -14376 (329)\tabularnewline
\hline 
iFHMM/7 & -13471 (274) & -14687 (385)\tabularnewline
\hline 
iFHMM/10 & -13949 (214) & -15116 (302)\tabularnewline
\hline 
iFHMM/15 & -15514 (201) & -16895 (342)\tabularnewline
\hline 
\end{tabular}
\par\end{centering}

\protect\caption{Training and Testing Data Log-likelihoods}
\end{table}

\begin{figure}[ht]

\begin{centering}
\label{K-fig} \includegraphics[bb=100bp 230bp 500bp 550bp,clip,scale=0.5]{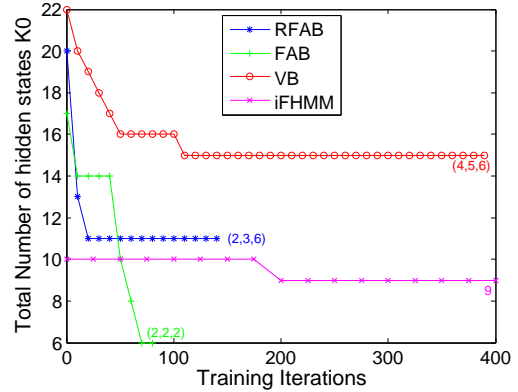}\vspace{-10pt}
\par\end{centering}

\centering{}\protect\caption{State Number Evolves over Iterations}
\end{figure}


Fig.2 shows how the hidden state numbers $K$ change over iterations
in a typical trial. The final state numbers at convergence are shown
at the end of each line. RFAB and FAB converge much faster than other
methods ($<200$ iterations), which shows the shrinkage regularization
accelerates the pruning of redundant states.

\begin{figure}
\begin{centering}
\includegraphics[bb=90bp 240bp 500bp 550bp,clip,scale=0.5]{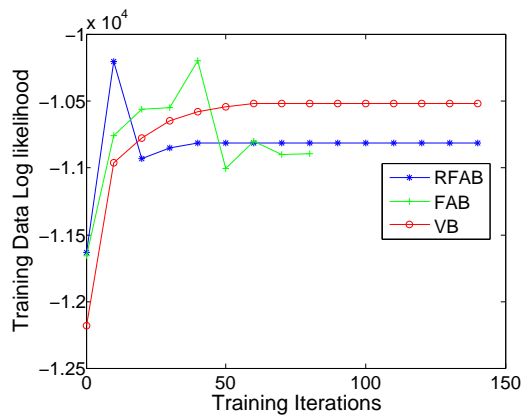}
\par\end{centering}

\protect\caption{Training Data Log-likelihood}
\end{figure}

Fig.3 shows how training data log-likelihood changes over iterations
in the same trial as in Fig.2. The log-likelihoods of RFAB and FAB
on the training data are quite close. The log-likelihoods of iFHMM
are not shown in this figure, as they are well below other methods
(around -13500).

\section{Conclusions and Future Work}

The FHMM is a flexible and expressive model, but it also poses a difficult
challenge on how to set its free, structural parameters. In this paper,
we have successfully extended the recently-developed FAB framework
onto the FHMM to address its model selection problem. We have derived
a better asymptotic approximation of the data marginal likelihood
than conventional FAB, by integrating out the initial and transition
probabilities. Based on this refined marginal likelihood, an EM-like
iterative optimization procedure, namely $\text{FAB}_{fhmm}$, has
been developed, which can find both good model structures and model
parameters at the same time. Experiments on a synthetic data set have
shown that $\text{FAB}_{fhmm}$ obtains more parsimonious and better-fit
models than the state-of-the-art nonparametric iFHMM and variational
FHMM.

In addition, we have proved that, during the $\text{FAB}_{fhmm}$'s
shrinkage process on hidden variables, if there are two very similar
hidden states, then one state would almost surely ``die out'' and
be pruned. This proof theoretically consolidates the shrinkage process
which we observe in experiments.

\bibliographystyle{abbrv}
\bibliography{fabfhmm}

\appendix
\clearpage

\section{Appendix}

\setcounter{thm}{0}
\begin{lem}
Suppose $\{z_{n,1},\cdots,z_{n,T_{n}}\}_{n=1}^{N}$ are $N$ sequences
of Bernoulli random variables, whose means are $\{p_{n,1},\cdots,p_{n,T_{n}}\}_{n=1}^{N}$.
For the $n$-th sequence $\{z_{n,t}\}$, $z_{n,1},\cdots,z_{n,T_{n}}$
are independent with each other. Let $y_{n}=\sum_{i}z_{n,i}$, $\bar{y}_{n}=E[y_{n}]=\sum_{i}p_{n,i}$.
Suppose further that $\bar{y}_{n}\to\infty$ as $T_{n}\to\infty$.
Besides, there are $N$ numbers $\{\hat{y}_{n}\}$, $\forall n,\hat{y}_{n}\approx\bar{y}_{n}$.
When all $T_{n}$ are large enough, the following bounds hold:
\end{lem}
1)
\begin{flalign*}
\mathbf{E}[\log(y_{n}+1)] & =\log(\hat{y}_{n}+1)+\frac{\bar{y}_{n}-\hat{y}_{n}}{\hat{y}_{n}+1}+\epsilon_{1};
\end{flalign*}

2)
\[
\mathbf{E}[y_{n}\log y_{n}]=\bar{y}_{n}\log\hat{y}_{n}+(\bar{y}_{n}-\hat{y}_{n})+\epsilon_{2}.
\]

\textit{Here $\epsilon_{1},\epsilon_{2}$ are small bounded errors.}

\vspace{10pt}

\noindent \textbf{Proof.}

1) Using the convexity of $-\log(y_{n}+1)$, we easily obtain

\begin{equation}
\mathbf{E}[\log(y_{n}+1)]\le\log(\mathbf{E}[y_{n}]+1)=\log(\bar{y}_{n}+1).\label{eq:logyupperbound}
\end{equation}

We proceed to derive a lower bound of $\mathbf{E}[\log(y_{n}+1)]$.

The following lower bound of the logarithm function is well know:

\[
\log x>1-\frac{1}{x}\quad\textnormal{for all }x>0.
\]

Substituting $x$ with ${\displaystyle \frac{y_{n}+1}{\bar{y}_{n}+1}}$,
we have

\[
\log(y_{n}+1)-\log(\bar{y}_{n}+1)>1-\frac{\bar{y}_{n}+1}{y_{n}+1}.
\]

After taking the expectation of both sides, it becomes
\begin{equation}
\mathbf{E}[\log(y_{n}+1)]-\log(\bar{y}_{n}+1)>1-(\bar{y}_{n}+1)\mathbf{E}\left[\frac{1}{y_{n}+1}\right].\label{eq:logyerror}
\end{equation}

$\mathbf{E}\left[\frac{1}{y_{n}+1}\right]$ is commonly referred to
as the \textit{negative moment} of $y_{n}$ \cite{negmoment}. We
apply the corollary in \cite{negmoment} that 

\begin{equation}
\mathbf{E}\left[\frac{1}{y_{n}+1}\right]=\int_{0}^{1}G_{n}(t)dt,\label{eq:negmoment}
\end{equation}
where $G_{n}(t)$ is the probability generating function of $y_{n}$.
It is known that $G_{n}(t)=\prod_{i=1}^{T_{n}}(q_{n,i}+p_{n,i}\cdot t)$,
in which $q_{n,i}=1-p_{n,i}$. Thus \eqref{eq:negmoment} becomes

\begin{equation}
\mathbf{E}\left[\frac{1}{y_{n}+1}\right]=\int_{0}^{1}\prod_{i=1}^{T_{n}}(q_{n,i}+p_{n,i}\cdot t)dt.\label{eq:negmoment2}
\end{equation}

We apply the \textit{Inequality of arithmetic and geometric means}
to $\prod_{i=1}^{T_{n}}(q_{n,i}+p_{n,i}\cdot t)$:

\begin{align*}
\textnormal{\ensuremath{\forall}}t\ge0:\quad\prod_{i=1}^{T_{n}}(q_{n,i}+p_{n,i}\cdot t) & \le\left(\frac{\sum_{i}q_{n,i}}{T_{n}}+\frac{\sum_{i}p_{n,i}}{T_{n}}t\right)^{T_{n}}\\
 & =(1-\frac{\bar{y}_{n}}{T_{n}}+\frac{\bar{y}_{n}}{T_{n}}t)^{T_{n}}.
\end{align*}

Let $\bar{p}_{n}$ denote $\frac{\bar{y}_{n}}{T_{n}}$, and $\bar{q}_{n}$
denote $1-\frac{\bar{y}_{n}}{T_{n}}$. Then \eqref{eq:negmoment2}
becomes 
\begin{equation}
\mathbf{E}\left[\frac{1}{y_{n}+1}\right]\le\int_{0}^{1}(\bar{q}_{n}+\bar{p}_{n}t)^{T_{n}}dt=\frac{1-\bar{q}_{n}^{T_{n}+1}}{(T_{n}+1)\bar{p}_{n}}<\frac{1}{\bar{y}_{n}}.\label{eq:negmomentbound}
\end{equation}

Plugging \eqref{eq:negmomentbound} into \eqref{eq:logyerror}, we
have

\[
\mathbf{E}[\log(y_{n}+1)]-\log(\bar{y}_{n}+1)>1-\frac{\bar{y}_{n}+1}{\bar{y}_{n}}=-\frac{1}{\bar{y}_{n}}.
\]

Thus $-\frac{1}{\bar{y}_{n}}$ is the lower bound of the approximation
error, which tends to 0 as $T_{n}\to\infty$. Hence 
\begin{equation}
\exists\epsilon_{1a}>0,s.t.\;\forall T_{n},\mathbf{E}[\log(y_{n}+1)]>\log(\bar{y}_{n}+1)-\epsilon_{1a}.\label{eq:logylowerbound}
\end{equation}

Combining \eqref{eq:logyupperbound} and \eqref{eq:logylowerbound},
we have

\begin{equation}
\big|\mathbf{E}[\log(y_{n}+1)]-\log(\bar{y}_{n}+1)\big|<\epsilon_{1a}.\label{eq:logyerrorbound}
\end{equation}

On the other hand, as $\hat{y}_{n}\approx\bar{y}_{n}$, the first
order approximation of $\log(\bar{y}_{n}+1)$ about $\hat{y}_{n}$
is good:

\begin{equation}
\log(\bar{y}_{n}+1)=\log(\hat{y}_{n}+1)+\frac{\bar{y}_{n}-\hat{y}_{n}}{\bar{y}_{n}+1}+\epsilon_{1b},\label{eq:yyhaterrorbound}
\end{equation}

where $\epsilon_{1b}$ is an error of order $o(\bar{y}_{n}^{-1})$
that tends to 0 when $T_{n}$ increases.

Combining \eqref{eq:logyerrorbound} and \eqref{eq:yyhaterrorbound},
we have

\[
\left|\mathbf{E}[\log(y_{n}+1)]-\log(\hat{y}_{n}+1)-\frac{\bar{y}_{n}-\hat{y}_{n}}{\bar{y}_{n}+1}\right|<|\epsilon_{1a}|+|\epsilon_{1b}|.
\]

In other words, $\log(\hat{y}_{n}+1)+\frac{\bar{y}_{n}-\hat{y}_{n}}{\bar{y}_{n}+1}$
approximates $\mathbf{E}[\log(y_{n}+1)]$ with a small bounded error.
In addition, this error tends to 0 quickly as the sequence length
$T_{n}$ increases.

2) It is easy to verify $y_{n}\log y_{n}$ is convex, and thus $ $
\begin{equation}
\mathbf{E}[y_{n}\log y_{n}]\ge\bar{y}_{n}\log\bar{y}_{n}.\label{eq:ylogyLB}
\end{equation}

We proceed to derive an upper bound of $\mathbf{E}[y_{n}\log y_{n}]$.

Applying the inequality $\log x\le x-1$ by substituting $x$ with
$\frac{y_{n}}{\bar{y}_{n}}$, we have

$ $
\begin{equation}
y_{n}\log\frac{y_{n}}{\bar{y}_{n}}\le y_{n}(\frac{y_{n}}{\bar{y}_{n}}-1)=\frac{y_{n}^{2}}{\bar{y}_{n}}-y_{n}.\label{eq:ylogyUB0}
\end{equation}

Taking the expectation of both sides of \eqref{eq:ylogyUB0}, we have

\begin{align}
\text{} & \mathbf{E}[y_{n}\log y_{n}]-\bar{y}_{n}\log\bar{y}_{n}\nonumber \\
\le & \frac{\mathbf{E}[y_{n}^{2}]}{\bar{y}_{n}}-\bar{y}_{n}\nonumber \\
= & \frac{\mathbf{Var}(y_{n})}{\bar{y}_{n}}\nonumber \\
= & \frac{\sum_{i}p_{n,i}^{2}}{\sum_{i}p_{n,i}}\le1.\label{eq:ylogyUB}
\end{align}

Combining \eqref{eq:ylogyLB} and \eqref{eq:ylogyUB}, we have

\begin{equation}
\mathbf{\text{\ensuremath{\big|}}E}[y_{n}\log y_{n}]-\bar{y}_{n}\log\bar{y}_{n}\big|\text{\ensuremath{\le}1}.\label{eq:ylogybound}
\end{equation}

Furthermore, when $\bar{y}_{n}\approx\hat{y}_{n}$, the first order
approximation of $\log\bar{y}_{n}$ about $\hat{y}_{n}$ is good,
i.e.:

\begin{align}
 & \bar{y}_{n}(\log\bar{y}_{n}-\log\hat{y}_{n})\nonumber \\
= & \bar{y}_{n}(\bar{y}_{n}-\hat{y}_{n})/\hat{y}_{n}+\epsilon_{2a}\nonumber \\
= & (\bar{y}_{n}-\hat{y}_{n})+\frac{(\bar{y}_{n}-\hat{y}_{n})^{2}}{\hat{y}_{n}}+\epsilon_{2a},\label{eq:ylogyerror}
\end{align}

where $\epsilon_{2a}$ is a small bounded error.

As $T_{n}\to\infty$ , both $\bar{y}_{n}$ and $\hat{y}_{n}$ tend
to $\infty$. Therefore $\frac{(\bar{y}_{n}-\hat{y}_{n})^{2}}{\hat{y}_{n}}$
is a small bounded error $\epsilon_{2b}$. Then \eqref{eq:ylogyerror}
becomes
\begin{equation}
\bar{y}_{n}\log\bar{y}_{n}-\bar{y}_{n}\log\hat{y}_{n}=(\bar{y}_{n}-\hat{y}_{n})+\epsilon_{2a}+\epsilon_{2b}.\label{eq:ylogyerror2}
\end{equation}

Combining \eqref{eq:ylogybound} and \eqref{eq:ylogyerror2}, we have
\[
\mathbf{E}[y_{n}\log y_{n}]=\bar{y}_{n}\log\hat{y}_{n}+(\bar{y}_{n}-\hat{y}_{n})+\epsilon_{2},
\]

where $|\epsilon_{2}|\le|\epsilon_{2a}|+|\epsilon_{2b}|+1$ is also
a small bounded error. 
\begin{cor}
$\{y_{n}\}$, $\{\bar{y}_{n}\}$ and $\{\hat{y}_{n}\}$ are defined
as the same as in \textup{Lemma 1}. Then the following bounds hold:
\end{cor}
\begin{enumerate}\renewcommand{\labelenumi}{\theenumi)}

\item
\begin{flalign*}
 & \mathbf{E}[\log\Gamma(y_{n})]\\
= & \bar{y}_{n}(\log\hat{y}_{n}-\frac{1}{2\hat{y}_{n}})-(\hat{y}_{n}+\half1\log\hat{y}_{n})\\
 & +\half1(\log2\pi+1)+\epsilon_{3};
\end{flalign*}

\noindent \item
\begin{align*}
 & \mathbf{E}\left[\sum_{n}\!\log\!\Gamma(y_{n})\!-\!\log\!\Gamma(\sum_{n}y_{n})\right]\\
= & \sum_{n}\bar{y}_{n}\!\left[\log\!\big(\frac{\hat{y}_{n}}{\sum_{m}\hat{y}_{m}}\big)+\frac{1}{2\sum_{m}\hat{y}_{m}}-\frac{1}{2\hat{y}_{n}}\right]\\
 & +\half1\log\big(\sum_{n}\hat{y}_{n}\big)-\half1\sum_{n}\log\hat{y}_{n}\\
 & +\half1(N-1)(\log2\pi+1)+\epsilon_{4}.
\end{align*}

\end{enumerate}

\textit{Here $\epsilon_{3},\epsilon_{4}$ are small bounded errors.}

\vspace{10pt}

\noindent \textbf{Proof.}

1) The Stirling's approximation for the log-Gamma function is: 
\begin{equation}
\log\Gamma(y_{n})=(y_{n}-\half1)\log y_{n}-y_{n}+\half1\log2\pi+\epsilon_{3a},\label{stirling}
\end{equation}

where $\epsilon_{3a}$ is an error term of $O(y^{-1})$, which goes
to zero quickly as $T_{n}$ increases. Therefore $\mathbf{E}[\epsilon_{3a}]$,
denoted as $\bar{\epsilon}_{3a}$, is also a small bounded number.

Taking the expectation of both sides of \eqref{stirling}, and applying
Lemma 1, we obtain 
\begin{align*}
 & \mathbf{E}[\log\Gamma(y_{n})]\\
= & \bar{y}_{n}(\log\hat{y}_{n}-\frac{1}{2\hat{y}_{n}})-(\hat{y}_{n}+\half1\log\hat{y}_{n})+\half1(\log2\pi+1)+\epsilon_{3}.
\end{align*}

2) Obtained by repeatedly applying 1), and combining terms involving
$\bar{y}_{n}$.
\end{document}